%% file: main.tex
\documentclass[10pt,twocolumn,letterpaper]{article}

\usepackage{cvpr}              

\input{preamble}

\usepackage[T1]{fontenc}

\definecolor{cvprblue}{rgb}{0.21,0.49,0.74}
\usepackage[pagebackref,breaklinks,colorlinks,citecolor=cvprblue]{hyperref}
\def\paperID{11705} 
\def\confName{CVPR}
\def\confYear{2024}

\title{NC-SDF: Enhancing Indoor Scene Reconstruction Using Neural SDFs with View-Dependent Normal Compensation}

\author{\text{Ziyi Chen}$^{1}$ \ \ \text{Xiaolong Wu}$^{2}$ \ \ \text{Yu Zhang}$^{1}$ \thanks{Corresponding author}
\\
$^{1}${\text{Zhejiang University}} \ \ $^{2}${\text{Amap Alibaba}}
}

\begin{document}

\maketitle
\input{sec/0_abstract}

\input{sec/1_introduction}

\input{sec/2_related_work}

\input{sec/3_method}

\input{sec/4_experiments}

\input{sec/5_conclusion}

\newpage 
{
    \small
    \bibliographystyle{ieeenat_fullname}
    \bibliography{main}
}
\end{document}

%% file: preamble.tex
\usepackage{multirow} 
\usepackage{tabularx}
\usepackage{makecell}
\usepackage{mathtools, nccmath} 
\usepackage{bm} 
\allowdisplaybreaks[2] 
\usepackage[dvipsnames]{xcolor}

\usepackage[symbol]{footmisc} 

\usepackage{tikz}
\newcommand*\circled[1]{\tikz[baseline=(char.base)]{
            \node[shape=circle,draw,inner sep=0.6pt] (char) {#1};}}


%% file: sec/0_abstract.tex
\begin{abstract}
State-of-the-art neural implicit surface representations have achieved impressive results in indoor scene reconstruction by incorporating monocular geometric priors as additional supervision. However, we have observed that multi-view inconsistency between such priors poses a challenge for high-quality reconstructions. In response, we present NC-SDF, a neural signed distance field (SDF) 3D reconstruction framework with view-dependent normal compensation (NC). Specifically, we integrate view-dependent biases in monocular normal priors into the neural implicit representation of the scene. By adaptively learning and correcting the biases, our NC-SDF effectively mitigates the adverse impact of inconsistent supervision, enhancing both the global consistency and local details in the reconstructions. To further refine the details, we introduce an informative pixel sampling strategy to pay more attention to intricate geometry with higher information content. Additionally, we design a hybrid geometry modeling approach to improve the neural implicit representation. Experiments on synthetic and real-world datasets demonstrate that NC-SDF outperforms existing approaches in terms of reconstruction quality.
\end{abstract}

%% file: sec/1_introduction.tex
\section{Introduction}
\label{sec:introduction}

\begin{figure}
    \centering
    \includegraphics[width=\linewidth]{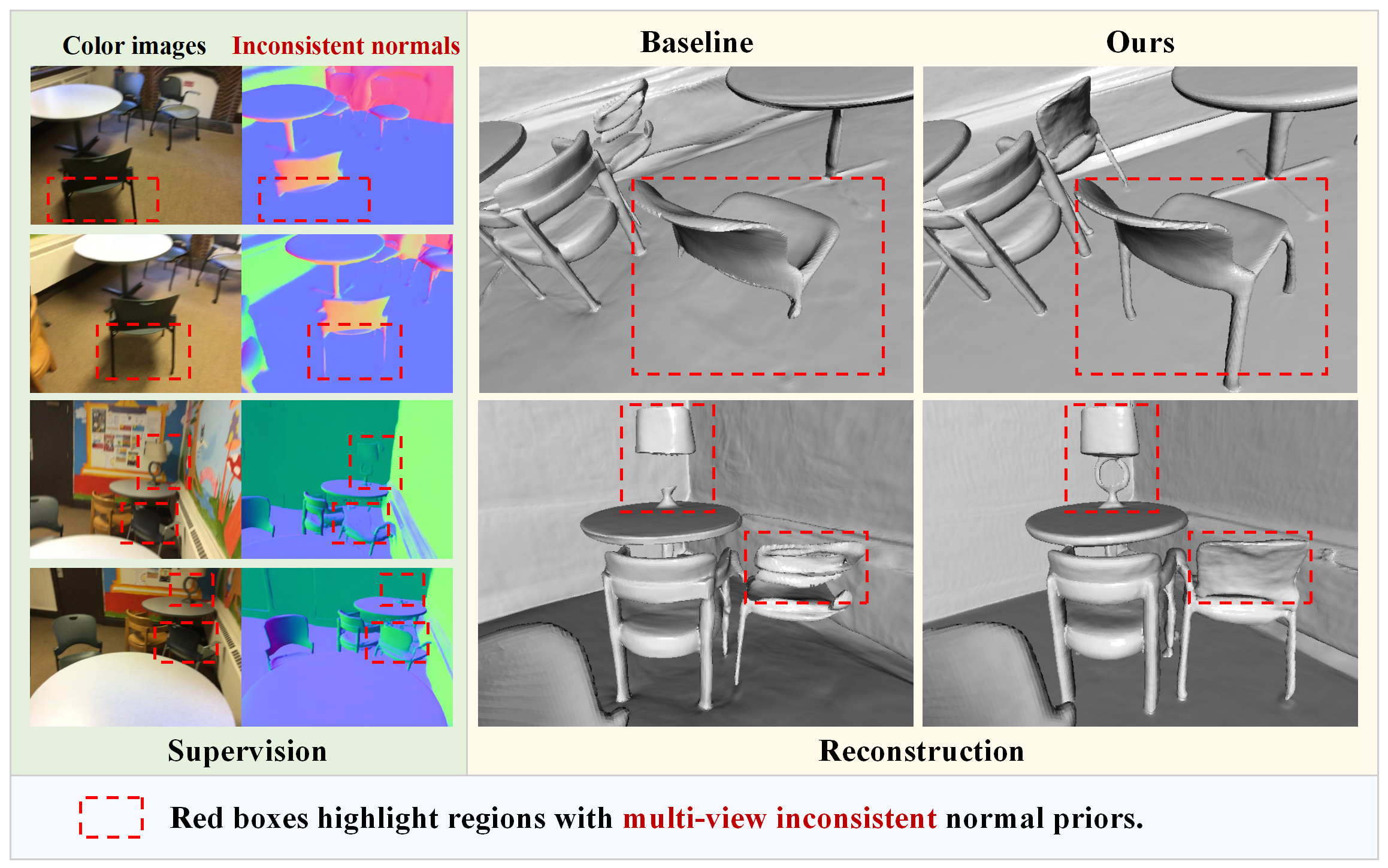}      
    \caption{\textbf{Comparison between baseline and NC-SDF.} State-of-the-art neural implicit surface representations produce suboptimal reconstructions with noisy or missing surfaces, primarily due to multi-view inconsistency between monocular geometric priors. Our NC-SDF introduces a view-dependent normal compensation model to adaptively learn and correct biases in normal priors. This approach enables the recovery of intricate geometric details while ensuring smoothness in texture-less areas within reconstructions.}
    \label{fig:intro}
\end{figure}


3D scene reconstruction from multi-view images is a long-term challenge in computer vision, with applications across various domains such as virtual reality, robotics, and autonomous driving. Multi-view stereo (MVS) techniques ~\cite{schonberger2016sfm, lindenberger2021pixelsfm, barnes2009patchmatch, bleyer2011patchmatch,yao2018mvsnet, huang2018deepmvs} recover depth maps for each view by matching features between adjacent views, subsequently fusing the depth maps to reconstruct 3D geometry. Their reconstructions tend to be noisy, especially in texture-less areas. Some data-driven methods ~\cite{murez2020atlas, sun2021neuralrecon,bozic2021transformerfusion, stier2021vortx, feng2023cvrecon} alleviate this limitation by directly predicting a truncated signed distance field (TSDF) from multi-view images. However, they necessitate expensive 3D supervision and produce over-smooth results.

Recently, impressive progress has been made in neural implicit surface reconstruction combined with volume rendering techniques. They utilize multi-layer perceptrons (MLPs) to parameterize implicit shape representations, such as occupancy~\cite{oechsle2021unisurf} or signed distance fields (SDFs)~\cite{wang2021neus, yariv2021volsdf}. Though these methods excel in capturing continuous and smooth surfaces, they face challenges when dealing with indoor scenes containing large texture-less regions. The primary reason is that multi-view photometric consistency fails to provide sufficient constraints in such regions, such as walls and floors.

Recent advancements have mitigated this problem by incorporating additional priors for supervision, including sensor depths~\cite{azinovic2022neuralrgbd, wang2022gosurf, ortiz2022isdf, zhu2022niceslam, zhang2023goslam, ruan2023slamesh}, semantic priors~\cite{guo2022manhattan}, depth priors from MVS methods~\cite{wei2021nerfingmvs, roessle2022dense, liang2023helixsurf} and monocular geometric priors~\cite{wang2022neuris, yu2022monosdf, zhu2023i2sdf, dong2023fastsdfrecon}. Among these works, ~\cite{wang2022neuris, yu2022monosdf} have produced state-of-the-art results by utilizing geometric cues from monocular geometry estimation networks~\cite{yin2019enforcing, wang2020vplnet, ramamonjisoa2019sharpnet, qi2018geonet, eftekhar2021omnidata}. However, their performance is heavily dependent on the quality of the geometric predictions. Notably, the estimation networks inevitably introduce biases between the predictions and the ground truths (GTs). Moreover, these biases are related to the viewing direction, since the networks receive input from a single view rather than multiple views. Therefore, such geometric predictions always struggle to satisfy multi-view consistency.

In this work, we present NC-SDF, a neural SDF 3D reconstruction framework with view-dependent normal compensation. The framework is designed to enhance indoor scene reconstruction by addressing multi-view inconsistency between monocular normal priors. To achieve this, We model not only the scene's radiance field and SDF but also the view-dependent biases in normal priors. Through adaptive compensation for the biases at corresponding viewing directions, our NC-SDF enables more consistent supervision and eventually leads to better performance. For more detailed reconstruction, we design an informative pixel sampling strategy to pay more attention to intricate geometry, by prioritizing sampling pixels with higher information content. Recognizing the limited representation power of MLPs, we introduce a hybrid geometry modeling approach based on feature fusion. This approach utilizes the inductive smoothness bias of MLPs to ensure smooth surfaces, and harnesses the high-frequency encodings provided by voxel grids to capture intricate geometry.

In summary, our NC-SDF significantly enhances the reconstruction quality. The combination of our three designs ensures consistent and smooth surfaces while enabling sharp details in the reconstructions. Our contributions can be summarized as follows:
\begin{itemize}
\item The view-dependent normal compensation model results in globally consistent and locally detailed reconstructions through adaptive compensation for the view-dependent normal biases.
\item The informative pixel sampling strategy and hybrid geometry model further enhance the reconstruction of geometric details.
\item Comprehensive experiments on both synthetic and real-world datasets demonstrate that our NC-SDF achieves state-of-the-art indoor scene reconstruction.
\end{itemize}

%% file: sec/2_related_work.tex
\section{Related work}
\label{sec:relatedwork}

\begin{figure*}
    \centering
    \includegraphics[width=\textwidth]{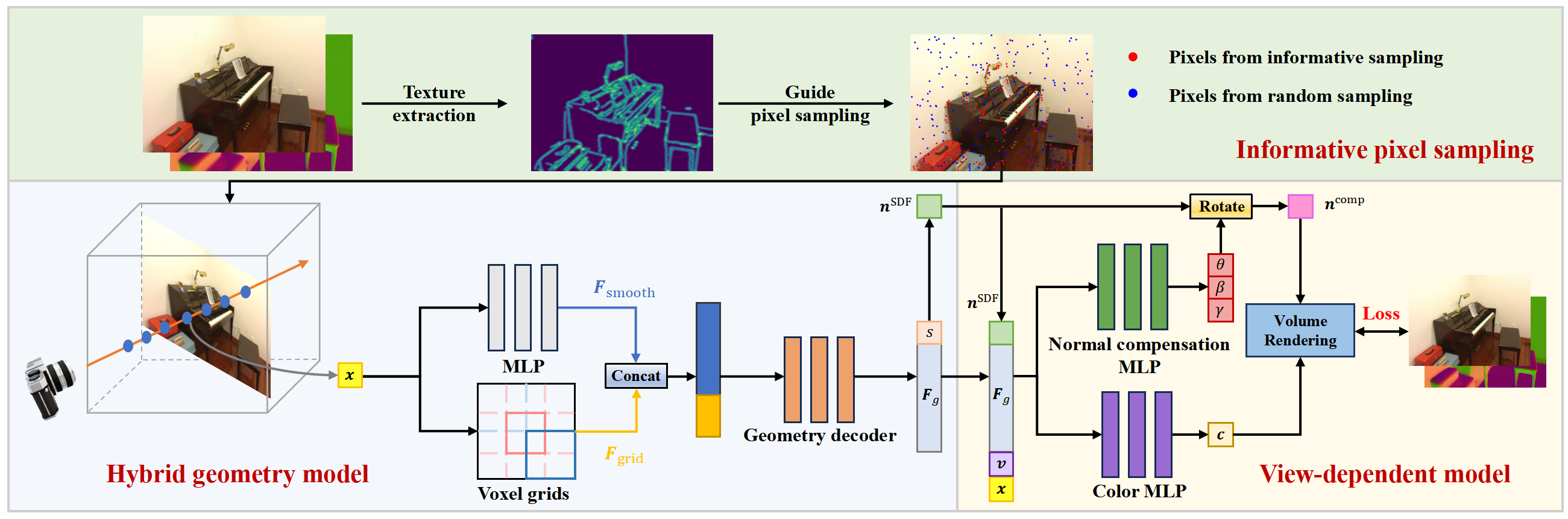}   
    \caption{\textbf{Pipeline of NC-SDF.} We model the geometry field (SDF), view-dependent radiance field, and view-dependent normal biases with neural implicit functions. Besides, we propose an informative pixel sampling strategy and a hybrid geometry model to further improve the reconstruction of thin geometry.}
    \label{fig:pipeline}
\end{figure*}

\subsection{Multi-view surface reconstruction}
\label{sec:2.1}
Traditional MVS methods~\cite{schonberger2016sfm, schonberger2016pixelwisemvs, bleyer2011patchmatch, barnes2009patchmatch, lindenberger2021pixelsfm} take multi-view images as inputs, and utilize feature matching and triangulation methods to estimate depth maps for each view. While excelling in reconstructing textured regions, they face difficulties when dealing with texture-less regions. With the development of deep neural networks, several learning-based MVS works~\cite{yao2018mvsnet, huang2018deepmvs, im2019dpsnet, yao2019recurrent, luo2019pmvsnet} utilize convolutional neural networks (CNNs) to learn the mapping from multi-view images to depth maps. However, the depth maps suffer from scale ambiguity and multi-view inconsistency. Alternatively, other learning-based methods~\cite{murez2020atlas, sun2021neuralrecon, feng2023cvrecon, bozic2021transformerfusion, stier2021vortx} directly predict the TSDF and then extract the mesh from the TSDF volume. These methods produce more consistent reconstructions, but they demand a large amount of ground truth 3D data for training, and the results often lack details due to the limitation of the TSDF resolution.

\subsection{Neural implicit surface reconstruction}
\label{sec:2.2}
Neural implicit functions have attracted increasing attention, owing to their advantages of compactness and low memory consumption. DeepSDF~\cite{park2019deepsdf} proposes to model the SDF of target objects with an MLP, achieving the reconstruction of complex shapes. Neural radiance fields (NeRF)~\cite{mildenhall2021nerf} and its variations~\cite{muller2022instant,fridovich2022plenoxels,barron2022mip,sun2022direct} utilize MLPs to implicitly encode the volume density and view-dependent emitted radiance field of the scene, enabling novel view synthesis. They optimize networks with only color images as constraints, by employing volume rendering techniques. Although iso-surfaces based on volume density can be extracted to recover the 3D geometry of the scene, the resulting mesh often exhibits noticeable noise due to the limited constraints on level sets.

To reconstruct smoother surfaces, several methods~\cite{wang2021neus,yariv2021volsdf} suggest using SDF as the output of neural implicit functions and reparameterizing SDF as volume density. This improvement results in superior surface reconstruction while preserving the capacity for novel view synthesis. However, due to the inductive smoothness bias of MLPs, using MLPs alone for scene modeling may result in over-smooth surfaces with limited details. Recent works~\cite{muller2022instant, zhu2022niceslam, wang2022gosurf, yu2022monosdf} combine voxel grids with a shallow MLP decoder to improve the representation power of the model. Despite improving the reconstruction of details, this approach introduces noise into the results because of under-constrained voxel grids.

\subsection{Prior-guided neural implicit surface reconstruction for indoor scenes}
\label{sec:2.3}
Constraints provided by color images are often insufficient for generating high-quality reconstructions when employing neural implicit representations in indoor scenes. This limitation primarily arises from the presence of large texture-less regions. Recent studies have introduced different kinds of priors as additional supervision to overcome the limitation. GO-Surf ~\cite{ wang2022gosurf} incorporates range measurements from depth cameras. Nerfingmvs~\cite{wei2021nerfingmvs} uses sparse depth information from Structure from Motion (SfM) to mitigate shape blurriness. ManhattanSDF~\cite{guo2022manhattan} assumes that the normals of walls and floors adhere to the Manhattan-world assumption. HelixSurf~\cite{liang2023helixsurf} utilizes MVS results to improve reconstruction quality and optimization time. NeuRIS~\cite{wang2022neuris} and MonoSDF~\cite{yu2022monosdf} explore monocular geometric cues from pretrained networks~\cite{yin2019enforcing, wang2020vplnet, ramamonjisoa2019sharpnet, qi2018geonet, eftekhar2021omnidata}. While they achieve state-of-the-art reconstruction results, their performance notably degrades in areas with inconsistent and noisy geometric priors. NeuRIS~\cite{wang2022neuris} filters out unreliable normal priors by checking multi-view photometric consistency during training, but the handcrafted strategy is vulnerable to noise in real-world datasets.

%% file: sec/3_method.tex
\section{Method}
\label{sec:method}

Given multi-view images with known poses, our goal is to produce high-quality 3D reconstructions. We begin with an introduction to our core framework and volume rendering technique (\cref{sec:3.1}). We then delve into our view-dependent normal compensation model (\cref{sec:3.2}), informative pixel sampling strategy (\cref{sec:3.3}), and hybrid geometry model (\cref{sec:3.4}). Finally, we provide details of loss functions (\cref{sec:3.5}). \cref{fig:pipeline} illustrates the pipeline of our NC-SDF.

\subsection{Preliminary}
\label{sec:3.1}

Our neural implicit representations consist of three components: the geometry model $f_g$, the color model $f_c$, and the normal compensation model $f_n$. In brief, the geometry model $f_g$ encodes the SDF, the color model $f_c$ encodes the view-dependent radiance field, and the normal compensation model $f_n$ encodes the view-dependent biases in monocular normal priors. We employ MLPs for the color and normal compensation modeling. And we utilize our proposed hybrid geometry model for the geometry modeling.

We adopt the differentiable volume rendering technique, following NeuS~\cite{wang2021neus}. A ray emitting from the camera center $\bm{o}$ can be expressed as $\bm{r} = \bm{o}\ + \ t\bm{v}$, where $\bm{v}$ represents the viewing direction of the ray. Along the ray, we sample $N$ points. For each 3D point $\bm{x}_i$, the geometry model $f_g$ maps it to a signed distance $s_i$ and geometry feature $\bm{F}_{g}$:
\begin{equation}
    s_i, \bm{F}_{g} = f_g(\bm{x}_i).
    \label{eq:fg}
\end{equation}

The color model $f_c$ outputs the radiance $\bm{c}_i$ observed from the viewing direction $\bm{v}$:
\begin{equation}
    \bm{c}_i = f_c(\bm{x}_i, \bm{v}, \bm{n}_i^\text{SDF}, \bm{F}_{g}),
    \label{eq:fc}
\end{equation}
where the normal $\bm{n}_i^\text{SDF}$ is the gradient of the signed distance $s_i$. In order to avoid confusion with other normals in subsequent discussions, we refer to this normal as the SDF normal. Then the color is accumulated along the ray:
\begin{equation}
    \hat{\mathbf{C}}=\sum_{i=1}^N T_i\alpha_i\bm{c}_i,
    \label{eq:render_c}
\end{equation}
where $T_i=\prod_{j=1}^{i-1}\left(1-\alpha_j\right)$ and $\alpha_i$ is the opaque density which can be further expressed as follows:
$$\alpha_i=\max \left(\frac{\Phi_\tau \left(s_i\right)-\Phi_\tau\left(s_{i+1}\right)}{\Phi_\tau\left(s_i\right)}, 0\right), \ \Phi_\tau(x)=(1+e^{-\tau x})^{-1}.$$

The normal compensation model $f_n$ predicts compensation rotation angles $\gamma, \beta, \theta$ about the $x, y, z$ axes from the viewing direction $\bm{v}$:

\begin{equation}
    \gamma, \beta, \theta = f_n(\bm{x}_i, \bm{v}, \bm{n}_i^\text{SDF}, \bm{F}_{g}).
    \label{eq:rotate}
\end{equation}

By applying the compensation rotation to the SDF normal $\bm{n}_i^\text{SDF}$, we obtain the compensated normal $\bm{n}_i^\text{comp}$. Further details are elaborated in \cref{sec:3.2}. We apply the same volume rendering technique to generate the rendered compensated normal map:

\begin{equation}
    {\mathbf{N}}^\text{comp}=\sum_{i=1}^N T_i\alpha_i\bm{n}_i^\text{comp}.
    \label{eq:render_n}
\end{equation}

We optimize the neural networks by minimizing the difference between the rendered outputs and reference inputs. The surface can be extracted as the zero level-set of the SDF using the Marching Cubes algorithm~\cite{lorensenmarchingcube}.

\subsection{View-dependent normal compensation model}
\label{sec:3.2}

As mentioned before, multi-view inconsistent priors can significantly impact reconstruction quality. ~\cite{roessle2022dense} utilizes per-view uncertainty maps from the prediction network to alleviate this problem, but these maps also fall short of satisfying multi-view consistency. Though NeuRIS~\cite{wang2022neuris} employs a geometric consistency checking strategy to filter out unreliable priors, the strategy requires manual threshold setting and lacks robustness. Inspired by volume rendering techniques~\cite{mildenhall2021nerf}, we propose integrating view-dependent biases in monocular normal priors into the implicit representation of the scene. Unlike previous methods, our approach compensates for the view-dependent biases in an adaptive way, without any additional manual operation.


\vspace{+1.5mm}
\noindent \textbf{Normal compensation.}\quad
The compensation process is as follows: we concatenate the spatial position of the point $\bm{x}$, its viewing direction $\bm{v}$, the SDF normal $\bm{n}^\text{SDF}$, and the geometry feature $\bm{F}_{g}$. This concatenated feature is then fed into our normal compensation model $f_n$, which outputs compensation rotation angles $\gamma$, $\beta$, and $\theta$ corresponding to the $x$, $y$, and $z$ axes, as described in \cref{eq:rotate}. Following \cref{eq:rotate_n}, we rotate the SDF normal $\bm{n}^\text{SDF}$ first by an angle $\gamma$ around the $x$-axis, then by an angle $\beta$ around the $y$-axis, and finally by an angle $\theta$ around the $z$-axis to obtain the compensated normal $\bm{n}^\text{comp}$. 

\begin{equation}
    \bm{n}^\text{comp} = \mathbf{R}_\text{ZYX} \bm{n}^\text{SDF} = \mathbf{R}_\text{Z}(\theta)  \mathbf{R}_\text{Y}(\beta)  \mathbf{R}_\text{X}(\gamma)  \bm{n}^\text{SDF},
    \label{eq:rotate_n}
\end{equation}
where $\mathbf{R}$ represents the corresponding rotation matrix, and the calculation is described in the supplementary material. We use the rendered compensated normal maps ${\mathbf{N}}^\text{comp}$ to align with the noisy normal priors, instead of directly supervising the rendered SDF normal maps $\mathbf{N}^\text{SDF}$. 

\vspace{+1.5mm}
\noindent \textbf{Two-stage training.}\quad
To stabilize the training, we design a two-stage training strategy. In the first stage, we optimize the color and geometry models to obtain a well-initialized radiance field and SDF. Previous studies~\cite{arpit2017closer, zhang2021understanding} have indicated that neural networks tend to memorize clean and easy patterns in the early stages of training. Therefore, our implicit functions can easily learn the distribution of regions with consistent supervision in the first stage. If training continues with inconsistent supervision, the neural network may eventually overfit on the noisy normals, leading to suboptimal reconstructions. To resolve this issue, we introduce the normal compensation model in the second stage and optimize it concurrently with the color and geometry models. By explicitly modeling the biases, our NC-SDF achieves more robust training under noisy supervision signals.

\begin{figure}[!t]
    \centering
    \setlength{\abovecaptionskip}{0cm}
    \begin{subfigure}[b]{0.45\textwidth}
      \centering
      \includegraphics[width=\linewidth]{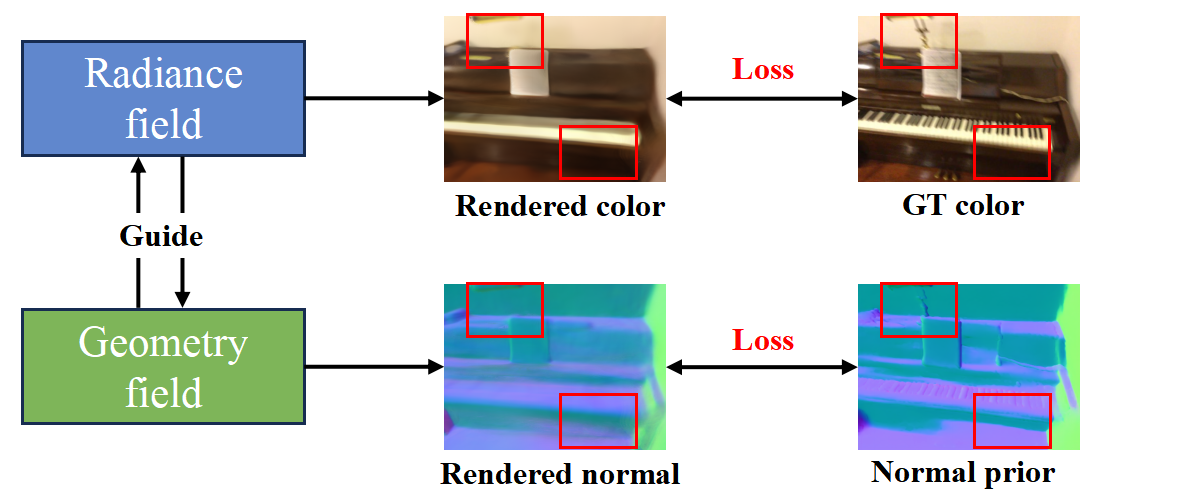}
      \caption{Without our normal compensation model.}
      \label{fig:sub_a}
    \end{subfigure}
    \vspace{0.5cm} 
    \begin{subfigure}[b]{0.45\textwidth}
      \centering
      \includegraphics[width=\linewidth]{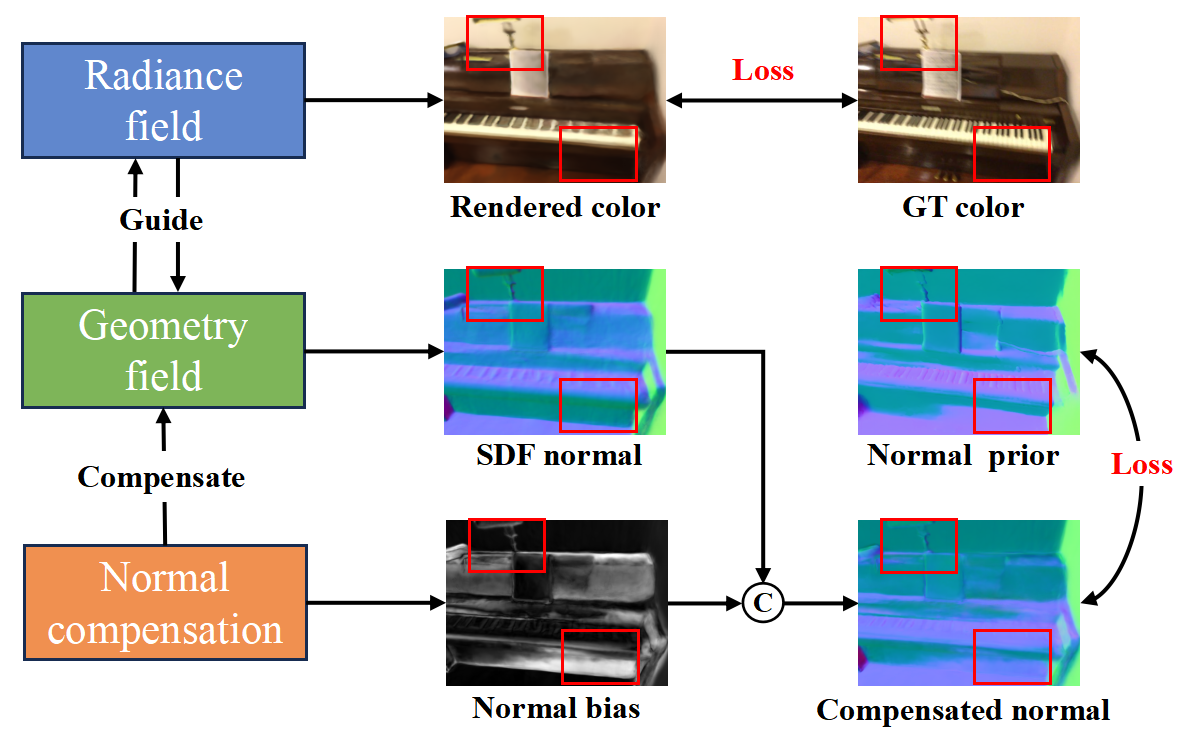}
      \caption{With our normal compensation model. \circled{\textbf{C}} represents the normal compensation process.}
      \vspace{-0.3cm} 
      \label{fig:sub_b}
    \end{subfigure}
    \caption{\textbf{Visualization of rendered results}, comparing (a) without and (b) with our normal compensation model.}
    \label{fig:normal_compensate}
\end{figure}

\cref{fig:sub_b} visually explains the working principle of the normal compensation (NC) model. The NC model's optimization direction is constrained by both RGB images and normal priors, rather than being arbitrary. Despite the noise in the normal priors, RGB images provide reliable supervision. In regions where normal priors exhibit multi-view inconsistency, color constraints play an essential role in facilitating the optimization of the geometry field. In summary, the coupled optimization relationship between the radiance field and the geometry field enables the NC model to disentangle normal priors reasonably.


\subsection{Informative pixel sampling strategy}
\label{sec:3.3}

In images of indoor scenes, texture-less regions usually occupy a significant portion of pixels, while geometric details are limited to a small portion. As previous research~\cite{he2009learning, krawczyk2016learning, sun2009classification} has pointed out, neural models exhibit biases towards the majority classes when trained on imbalanced datasets. Consequently, these models perform poorly on the minority classes. As is evident in neural implicit scene reconstruction, recovering intricate geometry proves to be more challenging than reconstructing plane surfaces. To address this issue, we propose an informative pixel sampling strategy as an alternative to the random pixel sampling method used in previous works~\cite{yariv2021volsdf, wang2021neus, wang2022neuris, yu2022monosdf}. 

\begin{figure}[t]
    \centering
    \includegraphics[width=\linewidth]{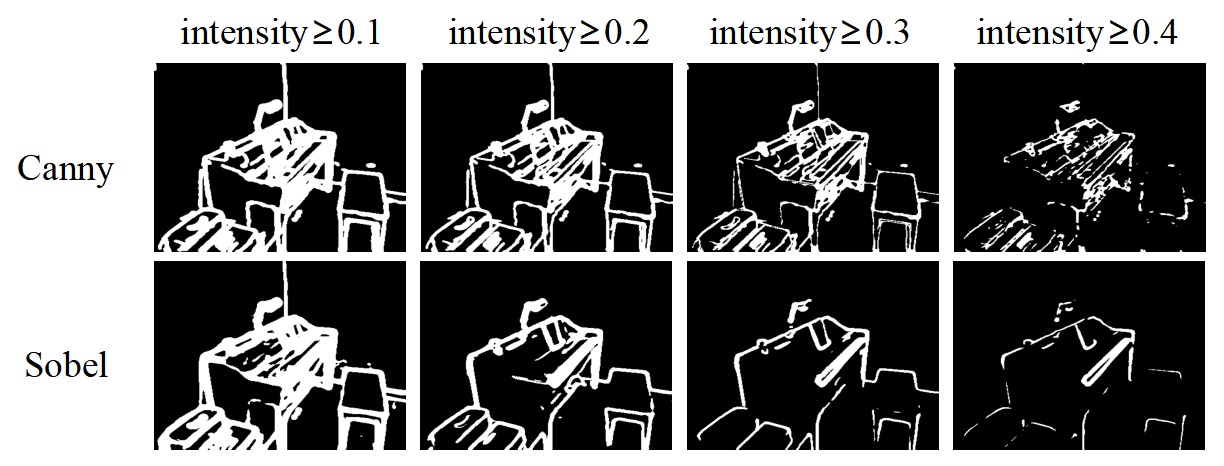} 
    \caption{\textbf{Visualization of texture intensity maps at different intensity thresholds}. The Canny operator exhibits a more robust performance than the Sobel operator.}
    \label{fig:canny_threshold}
\end{figure}

\vspace{+1.5mm}
\noindent \textbf{Texture extraction.}\quad
Image regions containing fine geometry often possess high information content, characterized by strong contrast and rich textures. We utilize the Canny edge detection operator~\cite{canny1983canny, ding2001canny} to capture these high-texture regions. The Canny operator is well-known for its robustness in extracting structural and textural information from visual objects. Compared to other gradient-based operators such as Sobel and Prewitt~\cite{shrivakshan2012comparison, ahmed2018comparative}, the Canny operator provides superior texture localization, noise reduction, and adaptability. In practice, we extract texture intensity maps from each image. The comparison in \cref{fig:canny_threshold} demonstrates that the Canny operator outperforms the Sobel operator in terms of robust texture extraction.


\vspace{+1.5mm}
\noindent \textbf{Pixel sampling strategy.}\quad
Our informative pixel sampling strategy evolves during training, adhering to a coarse-to-fine manner. We sample $N_\text{sample}$ pixels per batch. The pixel sampling is divided into two parts: a proportion $r$ is allocated to informative pixel sampling, while the remaining $1-r$ is allocated to random sampling. We set an intensity threshold $l_i$ for the extracted texture maps. From the set $\{l \ | \ l\geq l_i\}$, we randomly sample $r*N_\text{sample}$ pixels to create a high-information pixel set $\mathcal{P}_\text{canny}$. At the same time, we randomly sample $(1-r)*N_\text{sample}$ pixels from all the pixels, forming the set $\mathcal{P}_\text{random}$. Finally, the pixel set for each batch is expressed as $\mathcal{P}_\text{all} = \mathcal{P}_\text{canny} \cup \mathcal{P}_\text{random}$.

The proportion $r$ and intensity threshold $l_i$ evolve during training. Initially, we randomly sample pixels to establish a reliable initialization for the entire scene. As training progresses, our sampling strategy gradually transitions towards informative sampling, placing greater emphasis on intricate geometry. This coarse-to-fine approach enables more detailed reconstructions.

\subsection{Hybrid geometry model}
\label{sec:3.4}
Using MLPs alone for scene modeling is limited in capturing intricate details. Voxel grids provide improved expressive power, but their high dimensionality can introduce noise into reconstructions. To resolve this, we propose a hybrid geometry model based on feature fusion that combines the strengths of both MLPs and voxel grids models. The hybrid model comprises two feature branches: the MLP branch provides a smooth feature $\bm{F}_\text{smooth}$, encoding low-frequency structures, while the voxel grids branch offers a grid feature $\bm{F}_\text{grid}$, encoding high-frequency structures. We then concatenate $\bm{F}_\text{smooth}$ and $\bm{F}_\text{grid}$ to obtain a feature that encodes both low-frequency and high-frequency structures. The combined feature is subsequently decoded using a shallow MLP to obtain a geometry feature $\bm{F}_{g}$ and signed distance $s_i$. Our hybrid geometry model overcomes the limitations of using MLPs alone or voxel grids alone for scene modeling, striking a balance in modeling both intricate structures and planar regions.

\subsection{Loss functions}
\label{sec:3.5}
We use color images and normal priors for supervision. Furthermore, the gradients of the SDF satisfy the Eikonal equation~\cite{gropp2020eikonal}. We represent a set of camera rays passing through pixels as $\mathcal{R}$, color images as $\mathbf{C}$, and normal priors as $\mathbf{N}$. Our overall loss function is defined as follows:

\begin{equation}
    \mathcal{L}=\mathcal{L}_{c}+\lambda_n\mathcal{L}_{n}+\lambda_e\mathcal{L}_{e},
    \label{eq:loss}
\end{equation}
where
\begin{equation}
    \mathcal{L}_{c}=\sum_{\bm{r} \in \mathcal{R}}\|\hat{\mathbf{C}}(\bm{r})-\mathbf{C}(\bm{r})\|_1,
    \label{eq:loss_color}
\end{equation}

\begin{equation}
    \mathcal{L}_{n}=\sum_{\bm{r} \in \mathcal{R}}\|{\mathbf{N}^\text{comp}}(\bm{r})-{\mathbf{N}}(\bm{r})\|_1 +\left\|1-{\mathbf{N}^\text{comp}}(\bm{r})^{\text{T}}{\mathbf{N}}(\bm{r}) \right\|_1,
    \label{eq:loss_normal}
\end{equation}

\begin{equation}
    \mathcal{L}_{e}=\frac{1}{N}\sum_{\bm{r} \in \mathcal{R}}\sum_{i=1}^N\left(\left\|\nabla s_i\right\|_2-1\right)^2.
    \label{eq:loss_eik}
\end{equation}

%% file: sec/4_experiments.tex
\section{Experiments}
\label{sec:experiments}

\begin{figure*}[!htb]
    \centering 
    \includegraphics[width=\linewidth]{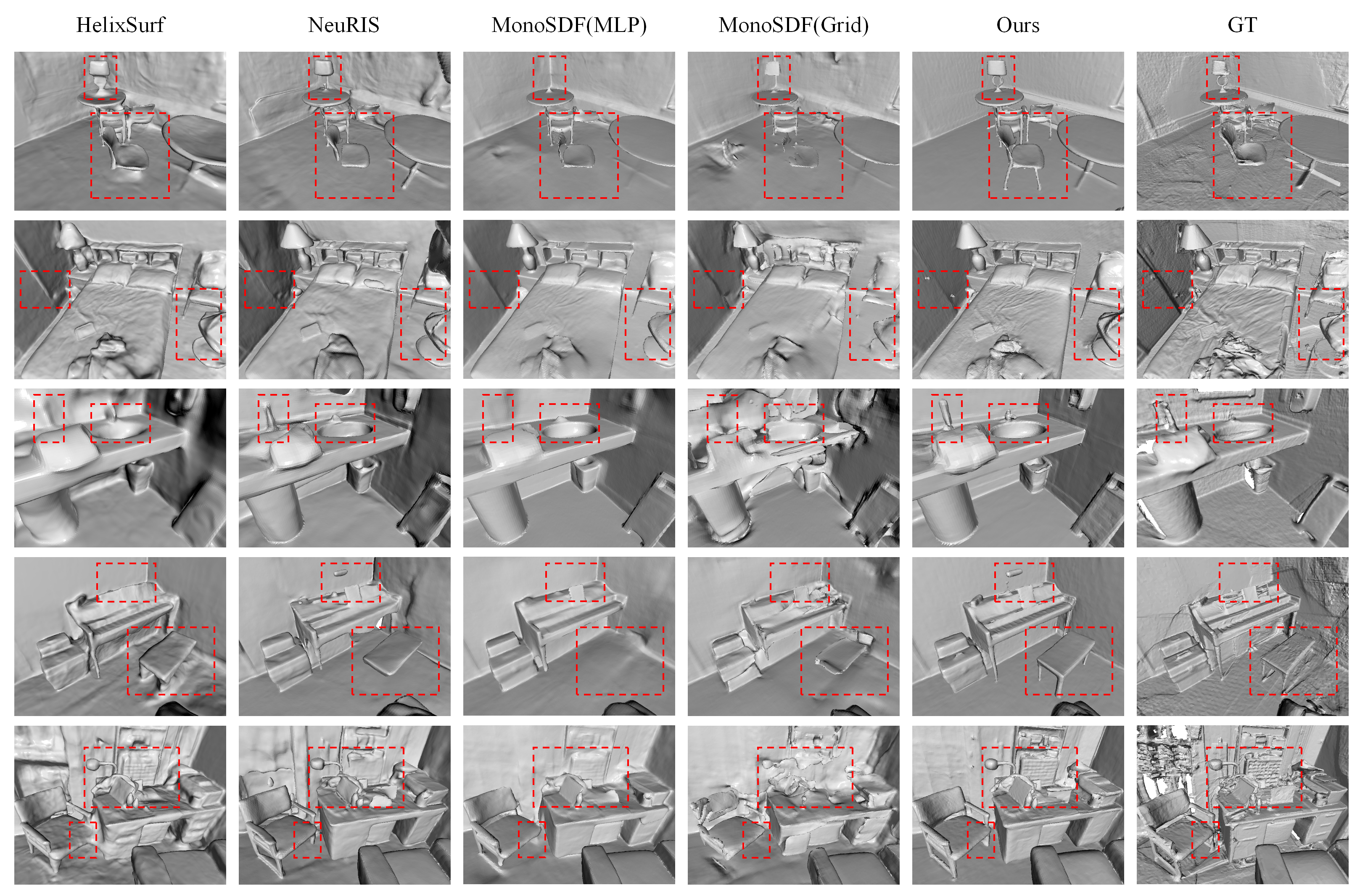} 
    \caption{Qualitative results on ScanNet~\cite{dai2017scannet}.}
    \vspace{-3mm} 
    \label{fig:scannet_contrast}
\end{figure*}

\subsection{Experiment Setup}
\label{sec:setup}

\vspace{+1.5mm}
\noindent \textbf{Datasets.}\quad
We choose two different indoor scene datasets for experiments, including ScanNet~\cite{dai2017scannet} and ICL-NUIM~\cite{handa2014benchmark}. These two datasets provide RGB-D images with camera poses. ScanNet is a real-world dataset and ICL-NUIM is a synthetic dataset. For comparison experiments, We select four scenes from each of the two datasets. For ablation studies, We use the four scenes from ScanNet. 

\vspace{+1.5mm}
\noindent \textbf{Baselines.}\quad
We compare our method with the following baselines: (1) Traditional MVS method COLMAP~\cite{schonberger2016sfm}; (2) Neural implicit representation methods without additional supervision, including VolSDF~\cite{yariv2021volsdf} and NeuS~\cite{wang2021neus}; (3) Neural implicit representation methods with additional supervision, including ManhattanSDF~\cite{guo2022manhattan}, HelixSurf~\cite{liang2023helixsurf}, NeuRIS~\cite{wang2022neuris}, and MonoSDF (both MLP and voxel grids version)~\cite{yu2022monosdf}.

\vspace{+1.5mm}
\noindent \textbf{Evaluation metrics.}\quad
Following \cite{murez2020atlas}, we evaluate the reconstruction results using accuracy, completeness, precision, recall, and F-score. The F-score is generally considered the most comprehensive indicator. 

\begin{table}[!htbp]
    \caption{Quantitative results on ScanNet~\cite{dai2017scannet}.}
    \resizebox{\linewidth}{!}{%
    \begin{tabular}{c|ccccc}
    \toprule 
    {Method}   &                     
    {Acc$\downarrow$} & 
    {Comp$\downarrow$} & 
    {Prec$\uparrow$} & 
    {Recall$\uparrow$} & 
    {F-score$\uparrow$}  \\
                            
    \midrule                        
    COLMAP~\cite{schonberger2016sfm} & 0.041 & 0.231 & 0.755 & 0.438 & 0.548  \\
    VolSDF~\cite{yariv2021volsdf} & 0.086 & 0.129 & 0.470 & 0.399 & 0.430  \\
    NeuS~\cite{wang2021neus} & 0.143 & 0.208 & 0.380 & 0.277 & 0.320  \\
    ManhattanSDF~\cite{guo2022manhattan} & 0.044 & 0.055 & 0.749 & 0.668 & 0.706  \\
    HelixSurf~\cite{liang2023helixsurf} & 0.036 & 0.042 & 0.791 & 0.725 & 0.756  \\  
    NeuRIS~\cite{wang2022neuris} & 0.051 & 0.050 & 0.709 & 0.662 & 0.684  \\
    MonoSDF(MLP)~\cite{yu2022monosdf} & 0.036 & 0.045 & 0.795 & 0.708 & 0.748  \\ 
    MonoSDF(Grid)~\cite{yu2022monosdf} & 0.048 & 0.050 & 0.727 & 0.663 & 0.693  \\ 
    Ours & $\mathbf{0.033}$ & $\mathbf{0.039}$ & $\mathbf{0.811}$ & $\mathbf{0.754}$ & $\mathbf{0.781}$  \\
    \bottomrule
    \end{tabular}
    }
    \label{table:quantitative_results_ScanNet}
\end{table}

\begin{figure}[!htb]
    \centering
    \includegraphics[width=\linewidth]{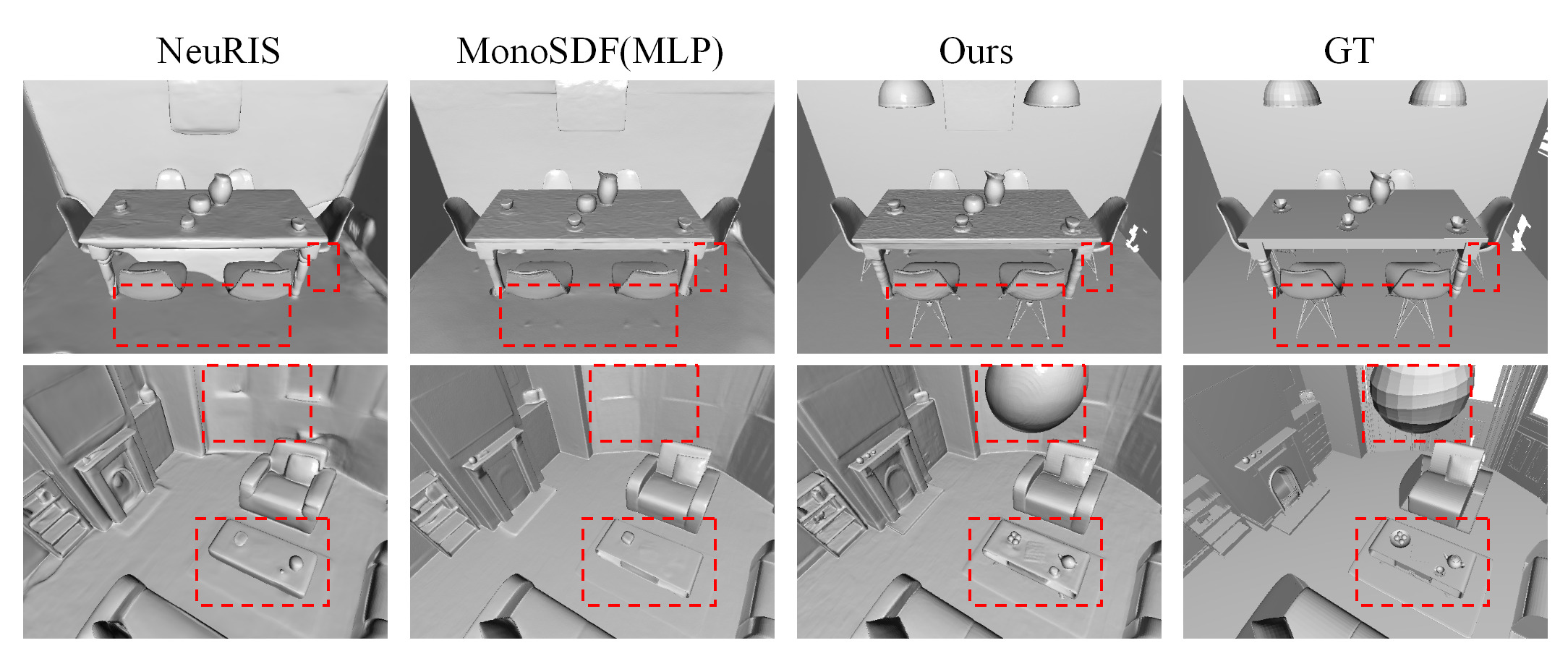}   
    \caption{Qualitative results on ICL-NUIM~\cite{handa2014benchmark}.}
    \label{fig:icl_contrast}
\end{figure}

\begin{table}[!htbp]
    \caption{Quantitative results on ICL-NUIM~\cite{handa2014benchmark}.}
    \resizebox{\linewidth}{!}{%
    \begin{tabular}{c|ccccc}
    \toprule 
    {Method}   &                
    {Acc$\downarrow$} & 
    {Comp$\downarrow$} & 
    {Prec$\uparrow$} & 
    {Recall$\uparrow$} & 
    {F-score$\uparrow$}  \\
                            
    \midrule      
    NeuRIS & 0.043 & 0.155 & 0.734 & 0.556 & 0.626  \\
    MonoSDF(MLP) & 0.043 & 0.143 & 0.741 & 0.589 & 0.650  \\ 
    Ours & $\mathbf{0.021}$ & $\mathbf{0.105}$ & $\mathbf{0.955}$ & $\mathbf{0.780}$ & $\mathbf{0.852}$  \\
    \bottomrule
    \end{tabular}
    }
    \label{table:quantitative_results_ICL}
\end{table}

\vspace{+1.5mm}
\noindent \textbf{Implementation.}\quad
All the experiments are conducted on one NVIDIA RTX 3090 GPU. We employ the OmniData model~\cite{eftekhar2021omnidata} to predict monocular normal priors, using images with a resolution of $384 \times 384$. The smooth feature branch of the geometry model uses an MLP with 4 hidden layers, while the grid feature branch utilizes 8-layers voxel grids, with each layer storing 4-channel features. The grids' resolution is adjusted according to the scene's complexity. The decoder of the geometry model, the color model, and the normal compensation model each is represented by an MLP with 4 hidden layers. We sample 1024 pixels per batch. The loss weights are set to $\lambda_n = \lambda_e = 0.1$. We use the Adam optimizer with an initial learning rate of $1 \times 10^{-3}$. The first training stage requires around 20,000 iterations. And the second training stage usually requires 60,000 to 80,000 iterations.

\begin{figure}[!t]
    \centering
    \includegraphics[width=\linewidth]{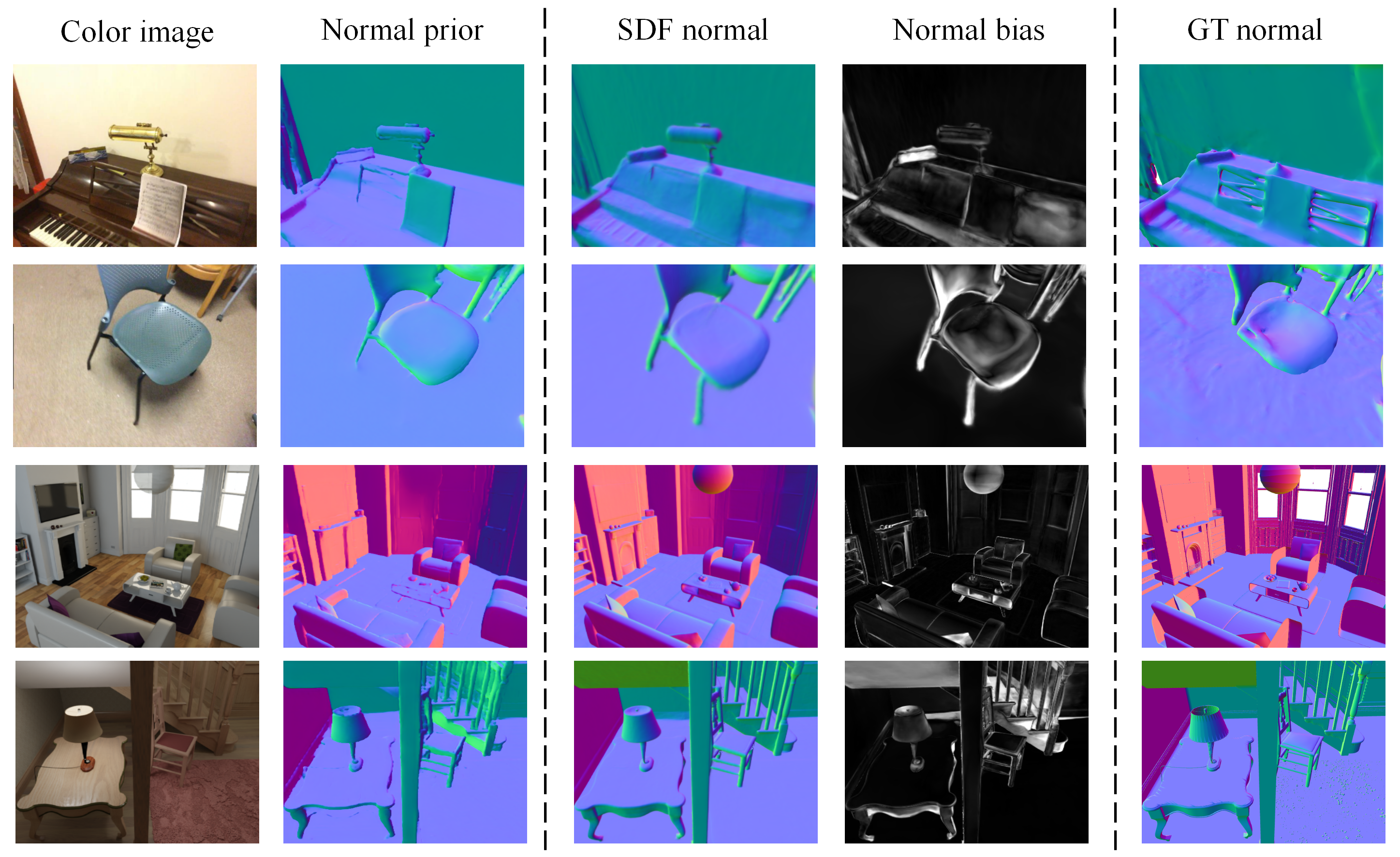}   
    \caption{Visualization of normal compensation.}
    \vspace{-1mm} 
    \label{fig:render_vis}
\end{figure}

\subsection{Comparisons}
\label{sec:comparisons}
\paragraph{Results in real-world dataset.}
\label{sec:real-world-dataset}
We visualize the reconstruction results of different methods in \cref{fig:scannet_contrast} and present the quantitative results in \cref{table:quantitative_results_ScanNet}. Both qualitative and quantitative results demonstrate that our method achieves the best reconstruction performance among all methods. COLMAP~\cite{schonberger2016sfm} and neural implicit representation methods without additional supervision including VolSDF~\cite{yariv2021volsdf} and NeuS~\cite{wang2021neus} produce unsatisfactory results in regions with weak textures. The methods with additional priors, such as ManhattanSDF~\cite{guo2022manhattan}, HelixSurf~\cite{liang2023helixsurf}, and MonoSDF (MLP)~\cite{yu2022monosdf} do improve the reconstruction quality; however, their reconstructions exhibit noisy or missing surfaces due to the limitations in the quality of priors and the expressiveness of the model. NeuRIS~\cite{wang2022neuris} designs a multi-view consistency checking strategy to filter unreliable normal priors. Despite contributing to more detailed reconstructions, the handcrafted strategy lacks robustness in handling real-world noise and finally leads to non-smooth surfaces. While MonoSDF (Grid) improves the reconstruction of fine geometry by enhancing the expressiveness of the geometry model, it generates noisy surfaces due to the lack of spatial consistency constraints for voxel grids. In contrast, our NC-SDF excels in capturing intricate geometry while producing smooth surfaces in texture-less regions.



\vspace{+1.5mm}
\noindent \textbf{Results in synthetic dataset.}\quad
Both the qualitative results in \cref{fig:icl_contrast} and the quantitative results in \cref{table:quantitative_results_ICL} validate that our NC-SDF significantly outperforms existing methods.

\begin{figure}[!t]
    \centering
    \includegraphics[width=\linewidth]{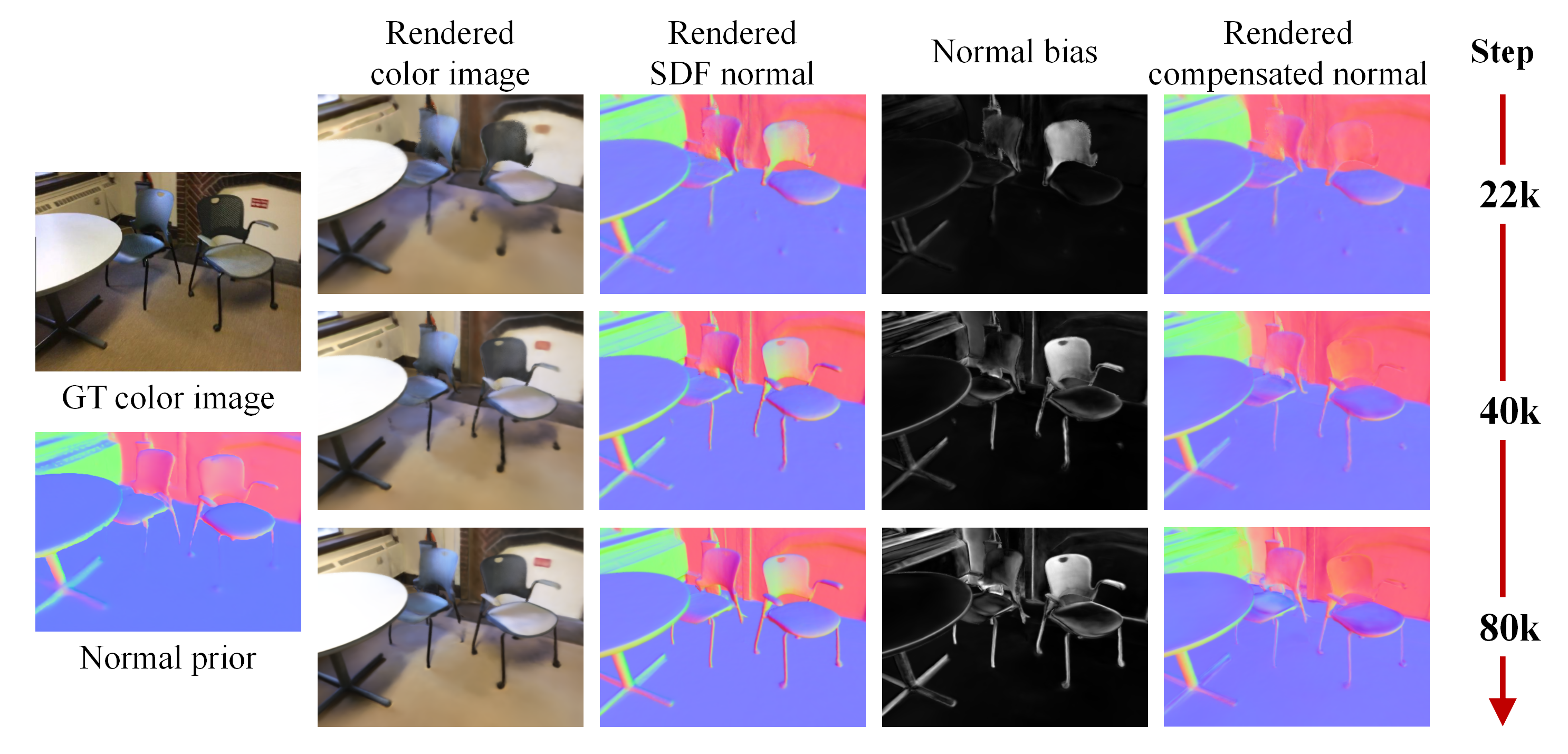}   
    \caption{Rendered results at different training stages. }
    \label{fig:render_results}
\end{figure}

Additionally, we visualize the rendered outputs related to the normal compensation in \cref{fig:render_vis}. The normal bias in the \cref{fig:render_vis} is computed as follows:
\vspace{-2mm}
\begin{equation}
    \mathbf{N}^\text{bias} = \sum_{j=1}^{3}|\mathbf{N}^\text{SDF}_j - \mathbf{N}^\text{comp}_j|,
    \label{eq:normal_bias}
\end{equation}
\vspace{-0.5mm}
where $\mathbf{N}^\text{SDF}_j$ and $\mathbf{N}^\text{comp}_j$ represent the rendered SDF normal map and the rendered compensated normal map, respectively, in the $j$-th channel. The visualization of the bias map proves that our NC model is capable of learning the biases in normal priors. Furthermore, we present the rendered results at different training stages in \cref{fig:render_results}. The results indicate that the NC model gradually learns the normal biases, thereby resulting in a gradual enhancement in the quality of both view synthesis and geometric reconstruction.

\begin{figure}[!b]
    \centering
    \includegraphics[width=\linewidth]{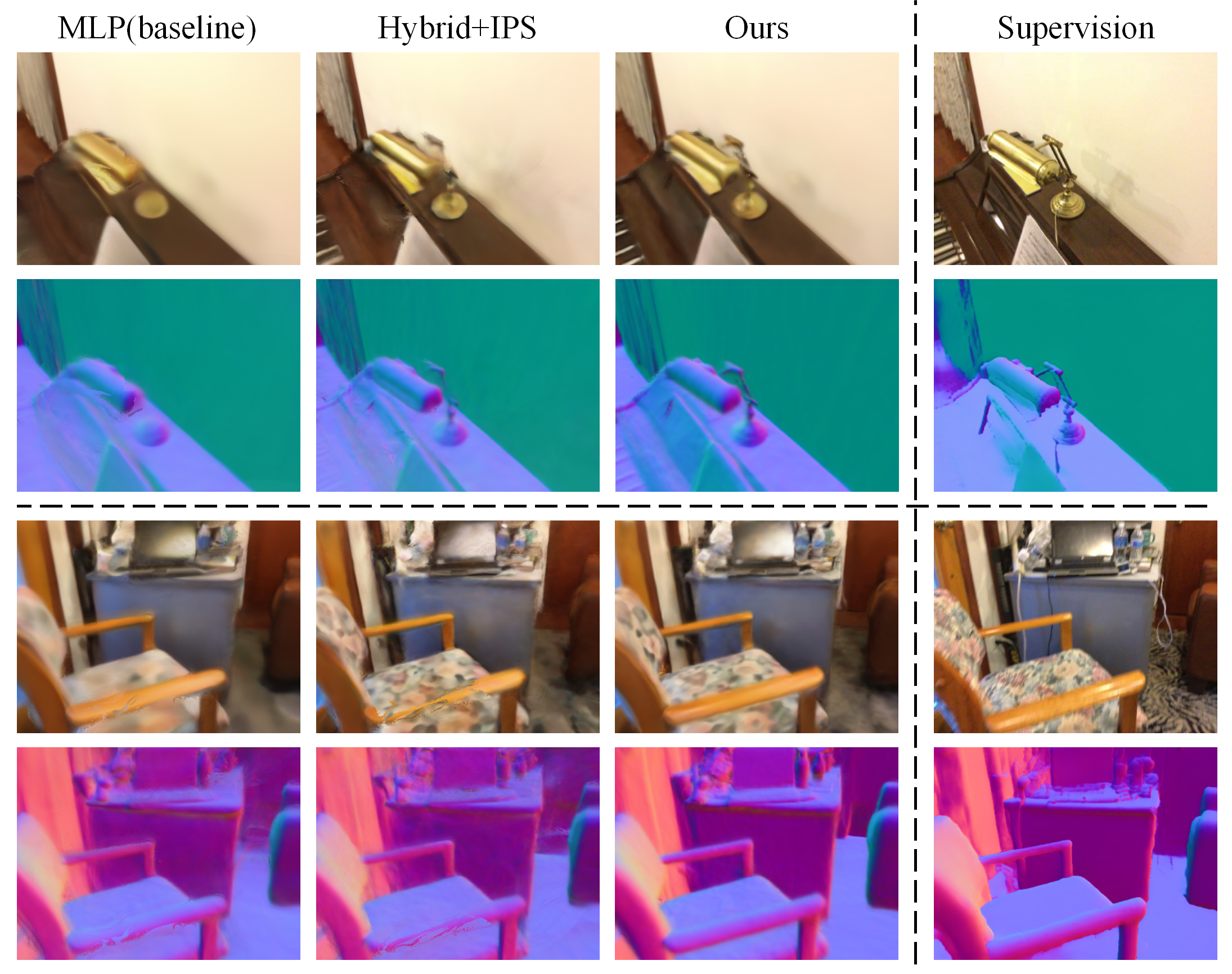}  
    \caption{Comparison of rendered results from three ablation experiments.}
    \label{fig:complement}
\end{figure}

\begin{figure*}[!htbp]
    \centering
    \includegraphics[width=\linewidth]{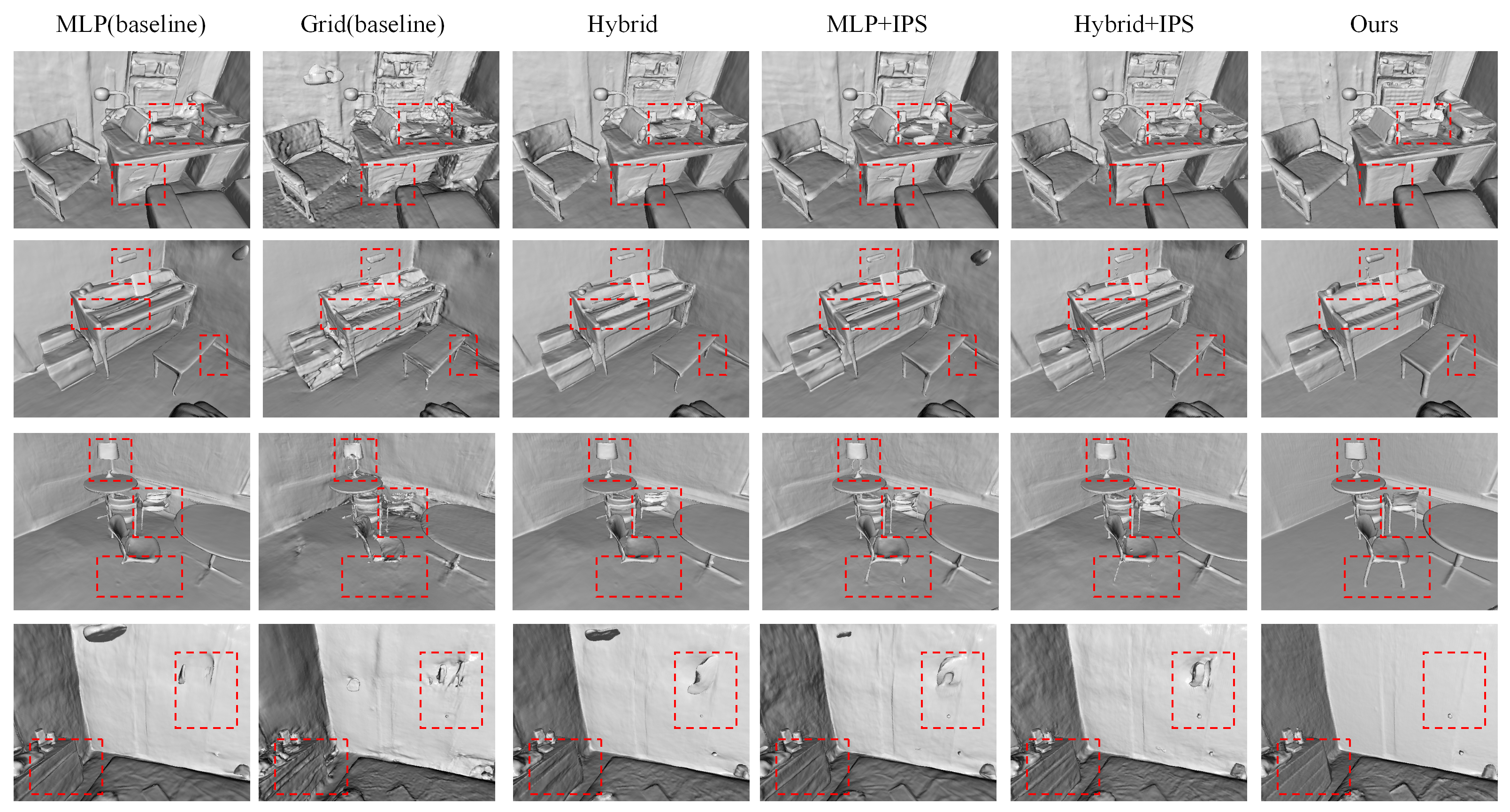}  
    \caption{Visualization of ablation studies on ScanNet.}
    \vspace{-3mm}
    \label{fig:scannet_ablation}
\end{figure*}


\begin{table*}[!htbp]
    \caption{Ablation studies on ScanNet.}
    \resizebox{\textwidth}{!}{%
    \begin{tabular}{c|cccc|ccccc}
    \toprule
    Method & MLP & Grid & \makecell{Informative \\ pixel sampling} & \makecell{Normal \\ compensation} & Acc$\downarrow$ & Comp$\downarrow$ & Prec$\uparrow$ & Recall$\uparrow$ & F-score$\uparrow$ \\ \midrule
    MLP(baseline) & \checkmark & & & & 0.040 & 0.042 & 0.750 & 0.718 & 0.733 \\
    Grid(baseline) & & \checkmark & & & 0.048 & 0.046 & 0.716 & 0.686 & 0.700 \\
    Hybrid & \checkmark & \checkmark & & & 0.039 & 0.041 & 0.766 & 0.727 & 0.745 \\
    MLP+IPS & \checkmark & & \checkmark & & 0.039 & 0.042 & 0.764 & 0.724 & 0.742  \\
    Hybrid+IPS & \checkmark & \checkmark & \checkmark & & 0.038 & 0.041 & 0.771 & 0.729 & 0.749 \\
    Ours & \checkmark & \checkmark & \checkmark & \checkmark & $\mathbf{0.033}$ & $\mathbf{0.039}$ & $\mathbf{0.811}$ & $\mathbf{0.754}$ & $\mathbf{0.781}$  \\
    \bottomrule
    \end{tabular}
    }
    \label{table:ablation}
    \vspace{-4mm}
\end{table*}
\vspace{-2mm}

\subsection{Ablation studies}
\label{sec:ablation}
For each of our three designs, we conduct the corresponding ablation studies. The quantitative results are shown in \cref{table:ablation} and the qualitative results are shown in \cref{fig:scannet_ablation}. The results indicate the effectiveness of each design, and the combination of these three designs yields the best performance.

We conduct experiments with six configurations: (1) \textbf{MLP(baseline)}: MLPs are utilized to model the SDF and radiance field. Random sampling is employed, and supervision is provided by normal priors and color images. (2) \textbf{Grid(baseline)}: A modification of (1) where voxel grids are used to model the SDF. (3) \textbf{Hybrid}: A modification of (1) where our hybrid geometry model is used to model the SDF. (4) \textbf{MLP+IPS}: A variation of (1) that integrates our informative pixel sampling (IPS). (5) \textbf{Hybrid+IPS}: This combines the hybrid geometry model with informative pixel sampling. (6) \textbf{Ours}: A combination of our three designs, including the informative pixel sampling, the hybrid geometry model, and the normal compensation model.

\vspace{+1.5mm}
\noindent \textbf{Effectiveness of the hybrid geometry model.}\quad
The comparison between \textbf{MLP(baseline)}, \textbf{Grid(baseline)}, and \textbf{Hybrid} illustrates that our hybrid geometry model enhances the reconstruction quality. \textbf{MLP(baseline)} tends to produce over-smooth surfaces while \textbf{Grid(baseline)} generates noisy surfaces. In contrast, \textbf{Hybrid} strikes a balance between the smoothness of surfaces and the sharpness of details.

\vspace{+1.5mm}
\noindent \textbf{Effectiveness of the informative pixel sampling.}\quad
Comparison between \textbf{MLP(baseline)} and \textbf{MLP+IPS}, as well as \textbf{Hybrid} and \textbf{Hybrid+IPS}, indicates that our sampling strategy results in a modest improvement in reconstruction quality. And it effectively enhances the reconstruction of geometric details, such as chair legs and doorknobs.



\vspace{+1.5mm}
\noindent \textbf{Effectiveness of the normal compensation model.}\quad
The comparison between \textbf{Hybrid+IPS} and \textbf{Ours} verifies that our normal compensation model alleviates the problems caused by multi-view inconsistency between monocular normal priors, including the non-smoothness of surfaces and the loss of details. \textbf{Hybrid+IPS} improves the reconstruction quality with a 2.18\% increase in F-score. The introduction of the normal compensation model further increases the F-score by 4.37\%. Furthermore, we visualize the rendered color images and the rendered SDF normal maps from three ablation experiments in \cref{fig:complement}. The comparison showcases that the normal compensation model yields a more accurate radiance field and geometry field.

%% file: sec/5_conclusion.tex
\vspace{-0.20cm}
\section{Conclusion}
\label{sec:conclusion}
\vspace{-0.15cm}

We present NC-SDF, a neural SDF 3D reconstruction framework with view-dependent normal compensation. The framework focuses on enhancing indoor scene reconstruction by addressing multi-view inconsistency between monocular normal priors. Specifically, we integrate view-dependent biases in normal priors into the neural implicit representation of the scene. In addition, we propose an informative pixel sampling strategy and a hybrid geometry modeling approach to further enhance reconstruction details. Experiments on real-world and synthetic datasets demonstrate that NC-SDF achieves state-of-the-art performance in indoor scene reconstruction. 

\noindent \textbf{Acknowledgement.}\quad 
This work was supported by STI 2030-Major Projects 2022ZD0208802, in part by NSFC 62088101 Autonomous Intelligent Unmanned Systems.

